\documentclass[submission,copyright,creativecommons]{eptcs}
\providecommand{\event}{SOS 2007} 
\usepackage{breakurl}             
\usepackage{underscore}   

\usepackage[dvipsnames]{xcolor}        

\usepackage{url}
\usepackage{graphicx}
\usepackage{amsmath}
\usepackage{amsthm}
\usepackage{booktabs}
\usepackage{algorithm}
\usepackage{algorithmic}
\newcommand{\be}{\begin{em}}
	\newcommand{\ee}{\end{em}}
\newcommand{\bb}{\begin{bf}}
	\newcommand{\eb}{\end{bf}}

\title{A Logic-based Multi-agent System for \\
	Ethical Monitoring and Evaluation of Dialogues}
\author{Abeer Dyoub
\institute{DISIM, University of L'Aquila, Italy}
\email{abeer.dyoub@univaq.it}
\and
Stefania Costantini
\institute{DISIM, University of L'Aquila, Italy}
\email{stefania.costantini@univaq.it}
\and
Ivan Letteri
\institute{DISIM, University of L'Aquila, Italy}
\email{ivan.letteri@univaq.it}
\and
Francesca A. Lisi
\institute{DIB \& CILA, University of Bari "Aldo Moro", Italy}
\email{FrancescaAlessandra.Lisi@uniba.it}
}
\def\titlerunning{A Logic-based Multi-agent System for
	Ethical Monitoring and Evaluation of Dialogues}
\def\authorrunning{A. Dyoub, S. Costantini, I. Letteri \& F.A. Lisi}
\begin{document}
	
	\maketitle
	
	\begin{abstract}
		Dialogue Systems are tools designed for various practical purposes concerning human-machine interaction. These systems should be built on ethical foundations because their behavior may heavily influence a user (think especially about children). The primary objective of this paper is to present the architecture and prototype implementation of a Multi Agent System (MAS) designed for ethical monitoring and evaluation of a dialogue system.
		A prototype application, for monitoring and evaluation of chatting agents' (human/artificial) ethical behavior in an online customer service chat point w.r.t their institution/company's codes of ethics and conduct, is developed and presented. Future work and open issues with this research are discussed.
	\end{abstract}
	
	\section{Introduction}
	
	Machine Ethics is an emerging field concerning itself with the ethical behavior of autonomous intelligent agents. Concerns about the ethical behavior of such machines is growing, especially with the increasing autonomy, and with agents 'invading' our everyday life and starting to perform many tasks on our behalf. 
	Engineering machine ethics, or building practical ethical machines is not just about traditional engineering. With machine ethics, we need to find ways to practically build machines that are ethically restricted, and can also reason about ethics. This involves philosophical aspects, even though the problem has a non-trivial computational nature.
	
	Chatbots are tools aimed at simplifying the interaction between humans and computers, typically used in dialogue systems for various practical purposes including customer service or information acquisition. From a technological point of view, a chatbot represents the natural evolution of question-answering system leveraging Natural Language Processing.
	Today, most chatbots are either accessed via virtual assistants such as Google Assistant and Amazon Alexa, or via messaging apps such as Facebook Messenger or WeChat, or via individual organizations' apps and websites.
	Business activities are rapidly moving towards the adoption of chatbots and other self-service technologies. This in order to automate basic communications and customer service, to reduce the call center costs and to provide advanced services to users. 
	However, chatbots raise many ethical concerns. Unethical Artificial Intelligence and bots are a big concern for many consumers. The chatbot should be built on ethical foundations because its behavior influences the company's image, and unethical behavior will lead to mistrust from the client-side.
	
	In previous works \cite{ADSCFL2019,ADILP2019,ADICLP2019}, a hybrid logic-based approach was proposed for ethical evaluation of chatbots' behavior, concerning online customer service chat points, w.r.t institution/company's codes of ethics and conduct. The approach is based on Answer Set Programming (ASP) as a knowledge representation and reasoning language \cite{gelfond2014knowledge}, and Inductive Logic Programming (ILP) for learning ASP rules needed for ethical evaluation and reasoning \cite{LawRB19}. The potential of logic-based approaches for programming machine ethics was discussed in \cite{abs-2009-11186}
	 
	In this paper, we focus on the challenge of monitoring and evaluating the ethical behavior of dialogue systems by proposing and implementing an application for monitoring and evaluation of chatting agents' (human/artificial) ethical behavior in an online customer service chat point w.r.t their institution/company's codes of ethics and conduct. The system is designed and implemented as a Multi-Agent System, and is based on the above mentioned ethical evaluation approach. The MAS acts as a separate ethical layer that can be integrated with existing dialogue systems. Our system is a pilot system aiming at first place to test the previously proposed ethical evaluation approach, and constitutes a step towards building practical ethical machines.
	

	
Our approach to the definition of an architecture for ethical monitoring and evaluation in dialogue systems is based on agent paradigm. The rationale behind this choice is our belief that the design of highly interactive ethical evaluators for dialogue systems requires higher level of software abstractions. Highly interactive environments, like in dialogue systems, require kinds of pro-activity and reactivity that objects and components (such as CORBA and COM-like components) as passive software entities are not able to exhibit. Moreover, agents are less dependent on other components than objects. 
	It is very easy to incorporate modifications in the behavior of individuals, by adding behavioral rules which act at the individual level. It is also possible to dynamically add new agents with their own behavioral model, which interact with the already defined agents, without having to recompile or even re-initiate the system.
Extensibility is one of the most powerful features of agent-based systems. The way in which agents are designed make them also easier to be reused than objects.

	Agent-oriented abstractions and multi-agent systems are well known in literature as  a programming  paradigm  for the  realization of complex and dynamic systems \cite{Jennings01}.  Accordingly, our implementation employs Logical Agents, and exploits relevant AOSE (Agent-Oriented Software Engineering) existing work. Namely, we adopt the JaCaMo \footnote{\url{http://jacamo.sourceforge.net}} methodology to design and implement a MAS simulation environment. 
	Via the JaCaMo platform, our application is realized by means of a set of Jason agents encapsulating the logic and the control of the specific tasks involved in the application, and operating with respect to the organizational constraints. Such constraints are defined through appropriate organizational artifacts which provide the functionalities, and the operations giving access to these functionalities, that agents can employ to perform their tasks in the specific context of the given application.
	
	
	The paper is organized as follows. We start, in Section~\ref{ethical} by recalling the ethical evaluation approach implemented in the proposed system. Section~\ref{arch} presents the proposed MAS  architecture. Then in the Section~\ref{eval}, an example showing the data flow through the system is presented. Finally, we conclude with discussion and future directions in Section~\ref{con}.  
	
	\section{Ethical Evaluation Approach}
	\label{ethical}

	The ethical evaluation approach implemented in the proposed system is based on previous work discussed in \cite{ADSCFL2019,ADILP2019,ADICLP2019}. This approach combines both top-down (rule-based) and bottom-up (learning) approaches in one unified hybrid framework. The approach is a purely declarative logic-based approach, that makes use of ASP as the main knowledge representation and reasoning language, and of ILP for learning the missing ASP rules needed for ethical reasoning. 
	The approach is based on the elaboration of facts extracted from documents containing the code of ethics and conduct that is proper of the given domain or organization, and from real life situations concerning pertinent ethical decision-making and judgment. These facts are used to elicit rules for ethical reasoning. The approach is general enough to produce ethical reasoning rules for any domain. 
	The motivation of devising such general approach is that
	Codes of Ethics in customer service are in general a set of abstract principles, aimed at objectively specifying the promises and obligations, related to the company’s products or services modalities of delivery, and to complaints management in the interaction with customers. For instance, such principles may include confidentiality, accountability, honesty, fidelity, etc. These ambiguous principles may carry different meanings according to contexts, and furthermore they are subject to interpretation. Therefore, it is quite difficult if not impossible to define such codes in a manner that they may be applied deductively. 
	And, is hardly possible for experts to define intermediate rules to cover all possible situations to which a particular code applies. In addition, there are many situations in which obligations might conflict.
	In \cite{ADSCFL2019,ADILP2019,ADICLP2019} we proposed an approach for generating the missing ethical detailed rules needed for ethical decision making and judgment via learning from interactions with customers over time. In our approach, the ethical evaluation agent will initially have in its knowledge base the set of ethical codes that provide a clear decision procedure which is encoded deductively using ASP. When the ethical evaluation agent does not have the proper rule to be able to provide an ethical evaluation of a certain case scenario, the needed rule will be learned by means of the learning module which uses ILP for this purpose.
	
	\section{EthicalEvalMAS: Architecture}
	\label{arch}
	\begin{figure}[!h]
		\centering
		\includegraphics[scale=0.4]{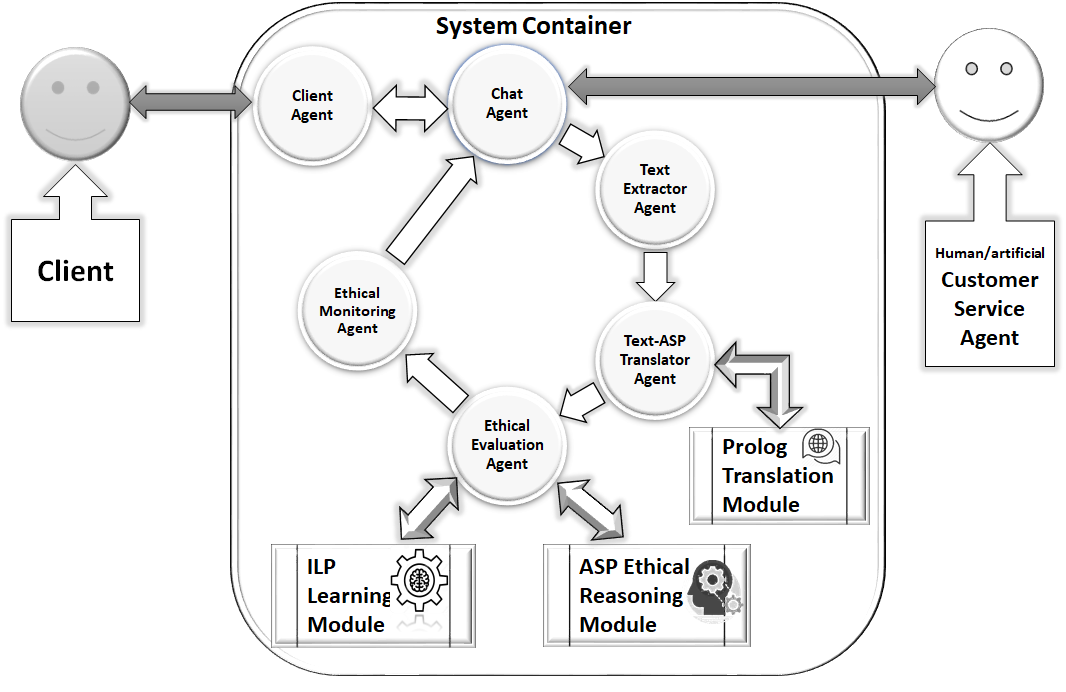}
		\caption{EthicalEvalMAS Architecture}\label{fig:architect}
	\end{figure}
	
	In this section we present the proposed MAS architecture. In the context of an online customer service dialogue system, we want to ensure an ethical behavior from the chatting agent (human/artificial). Online customer service agents are monitored for ethical violations by the proposed architecture. In order to achieve this overall goal, the MAS is composed of a group of agents, each one is responsible for a specific sub-task in the overall ethical monitoring task. The architecture is shown in Figure~\ref{fig:architect}. 
	The online customer service environment in this work consists of clients, online customer service agents (human/artificial), and software agents.
	A client interacts with the system via chatting point interface, where she/he can write their requests (questions), and receive answers. Answers to the client's requests are given by the online customer service agent. Software agents in the environment are: client agent (CA), chatting agent (ChA), text extractor agent (TEA), text-ASP translation agent (TATA), ethical evaluation agent (EEA), and monitoring agent (MA). Text extractor agent is responsible for extracting chat text from the chatting point interface and sending it to the TATA agent which is responsible for translating the chat text into ASP syntax. This agent uses Prolog\footnote{\url{https://sicstus.sics.se/}} module for this purpose (a simple translator was written in Prolog, this translator is able to translate simple natural language sentences into Prolog-like predicates). The ethical evaluation agent has two primary goals: (1) to generate ethical evaluation of the online customer service agent's answers using ASP reasoning module, which utilizes the current case facts, and the background knowledge (BK) from the knowledge base (KB) to give the evaluation. This module is based on Clingo solver\footnote{\url{https://potassco.org/clingo/}}. (2) Learning the ethical rules needed for ethical evaluation, saving it to its KB, in case the ASP reasoning module is not able to give an evaluation (i.e. no answer set). EEA uses the ILP learning module to achieve this purpose. Finally, the MA agent currently responsible only for alerting the CA agent for ethical violations (the role of this agent can be extended to practice more control over the CA agent).  
	
	For building our MAS model, we have used the JaCaMo framework. This framework allows us to program our MAS in terms of an organization of cognitive agents, sharing a common artifact-based environment. The developed application is available on github\footnote{\url{https://github.com/abeer-dyoub/EthicalEvalMAS}}.
	
	\section{Evaluation}
	\label{eval}

	The following briefly describes a scenario to demonstrate the usability of the system.
	
	\paragraph{Example scenario:} A client contacts an online customer service chat point asking about the characteristics of a certain product, and the dialogue system answers trying to convince the customer to buy the product. It starts saying that the product is environmentally friendly (which is irrelevant in this case), and that this is an advantage of their product over the same products of other companies. Such an answer, containing the use of irrelevant sensitive slogans to manipulate customers, is considered unethical.
	
	The process begins with the user entering the question: 'what are the features of ProductX?', through her/his chatting interface.
	The \textit{CA} gets the question text entered by the user through the chat point interface, and sends it in a message to the \textit{ChA}. The \textit{ChA} provides the answer:' ProductX is environmentally friendly', using its chatting interface, and this answer is sent to the user. 
	The \textit{TEA} focusing on the workspace of the \textit{ChA}, will extract the answer text from the chat point and will send it to the \textit{TATA}, which will show it via the environmental translation artifact, and will translate the composing sentences into ASP syntax (literal: {\small $environmentally\_friendly(productX)$}).
	The \textit{TATA} achieves this by means of a specific plan through which it invokes a particular operation, this operation invokes the \textit{prolog translation module}. this module will do the translation and return the result.
	The result of the translation in our case will be ASP predicates (facts). These facts are sent by the \textit{TATA} to the \textit{EEA}, and will be added to its knowledge base (KB). This agent has in her knowledge base the ontology of the domain including the following fact:	\textit{ {\small $sensitiveSlogan(environmentally\_friendly(productX))$}}, and the following ASP ethical evaluation rule (learned):
	\begin{multline*}
		\textit{{\small $unethical(V1) :- sensitiveSlogan(V1), not \hspace{1mm} relevant(V1)$}},
		\textit{{\small $answer(V1).$} }
	\end{multline*}
	The agent has no information about the relevance of the adoption of this sensitive slogan for the requested product, so it will safely assume by default the irrelevance. Then, the reasoner will infer the following evaluation as a result:
		{\small \textit{$unethical(environmentally\_friendly(productX)).$}}
	If subsequently we add to the KB of the \textit{EEA} the fact:
		{\small \textit{$relevant(environmentally\_friendly(productX))$}}. Then the ASP reasoner will no longer infer that the answer is unethical. 
	Once the \textit{EEA} receives the translation value from the \textit{TATA}, it will invoke a particular environmental operation which calls upon the \textit{ASP reasoning module}. This module will calculate a model for the above ASP program. If the model contains either of the literals $ethical(A)/unethical(A)$, then the corresponding environmental observable property will be updated, and perceived by the \textit{EEA}.
	The evaluation result along with the justification are shown through a specialized environmental artifact, and sent to the \textit{MA}, which will send a notification message to the employee agent (chatting agent).
	
	Now let us consider the situation before having the above mentioned rule for ethical evaluation in the \textit{EEA} knowledge base. The \textit{EEA} will not be able to give an ethical evaluation for the current case scenario, i.e. in the \textit{ASP reasoning module} output model there are non of the literals $ethical(A)/unethical(A)$, so the evaluation result is empty. At this point the \textit{EEA} will invoke the ILP learning module via a specialized GUI environmental artifact. This artifact will allow a human expert supervisor to label the case as ethical/unethical and give the needed information to be passed to the ILED learning algorithm \footnote{The ILED algorithm implementation which we have used can be found here: \url{https://github.com/nkatzz/ILED}} for learning the needed ASP ethical evaluation rule, then add it to the KB of the \textit{EEA}, after that signals the \textit{EEA} which will invoke again the ASP reasoning module to re-evaluate the current case scenario and produce the needed evaluation.

	So far, we have tested our prototype with a very small set of examples. However, our experiments are still limited due to the absence of a big enough dataset, which is one of the main challenges in the ethical domains in general (the lack of datasets and benchmarks was discussed lately at the AAAI 2021 Spring Symposium on Implementing AI Ethics). For this purpose, we developed and published a web application meant to collect data, regarding unethical scenarios, for producing a big training dataset to train our online customer service chatbot. The application is currently available online for participation\footnote{\url{http://ethicalchatbot.sytes.net/en/}}.  
	
	\section{Discussion, Conclusion, and Future Works}
	\label{con}
	This paper presented a multi-agent system architecture capable of ethical monitoring and evaluation of a dialogue system. A brief scenario was used to demonstrate the feasibility of the system. The developed MAS acts as a separate ethical component (ethical layer) for ethical evaluation, which provides many advantages from an engineering point of view: I) The ethical component has access to all data used for ethical evaluation, and use this data to provide justifications for a given ethical evaluation to humans, which leads to accountability. II) The possibility to adapt the ethical component to changes in circumstances and needs. In addition to, the possibility of implementing more than one version of the ethical component on the same agent. III) The possibility to check and verify the functionality of the ethical component independently from the operations of the autonomous agent. IV) The re-usability and standardization. Having a separate component for ethical evaluation gives us the possibility to standardize this ethical component, which will have the advantage of avoiding the need to re-invent ethical components that fit for a large number of agent' architectures.    
	
	The ethical evaluation of the proposed MAS system is based on the facts extracted from the case scenario, and their relation to the codes of ethics and conduct, which results in a set of ethical evaluation rules, against which to evaluate the behavior of the chatting agent. These rules are used to decide whether the chatting agent's answers to clients requests are ethical/unethical. Evaluating the decidability and completeness of the generated rules is an open issue, and is a matter of further experiments and evaluation.
	Evaluating the performance of moral machines is a hard task. So far,
	there is no objective and measurable criteria for the evaluation of moral
	machines. The complexity of this task comes from two questions: what
	exactly constitutes a moral execution of a task?; how to evaluate the moral
	behavior of a machine, i.e. against what?.
	The complexity of the ethical domain, from the ontological and epistemic
	point of view, renders the task of setting up an assessment criteria for ethical
	machine performance evaluation very hard. Considering our system and our
	application domain, there are objective facts as to whether a certain answer
	is considered ethical or not. These facts constrain whether a certain answer
	is correctly classified as ethical/unethical. We have supported our claim of
	the potential of our approach for creating ethical machines informally, by
	showing few example scenarios. However, the behavior of our ethical agent
	will be guided by the ethical theory generated by the system. To which level
	the generated theory can result in an acceptable ethical behavior, is related
	to the building process itself. In other words, it is extremely related to
	the correct specifications of what constitutes a moral behavior in a specific
	domain. A possible strategy that can help understand and define correct
	ethical specifications could be building an ontology for the domain’s ethics.
	From the technical and formal point of view, we can easily provide a logical
	proof that our system behaves correctly according to the specifications of
	moral customer service (once these specifications are defined). Given certain premises, the system can be proven to do what it should do. This proof can
	be done manually, or via theorem prover which uses automated logical and
	mathematical reasoning.
	
	Our System incorporates ASP as a non-monotonic knowledge representation and reasoning formalism, used for ethical reasoning via the ASP reasoning module. And ILP as a logic-based machine learning for learning logical rules for ethical reasoning via ILP learning module. This, in fact, increases the reasoning capability of our \textit{EEA} agent; promotes the adoption of hybrid strategies that allow both top-down design and bottom-up learning via context sensitive adaptation of models of ethical behavior; allows the generation of rules with valuable expressive and explanatory power,
	which equips our agents with the capacity to give an ethical evaluation, and
	explain the reasons behind this evaluation. In other words, this contributes to the transparency and accountability, which facilitates instilling confidence and trust in our agents.
	
	Providing explanations to system’s decisions is fundamentally linked to its reliability and trustworthiness. The ASP-program models contain both the output and the justification for the given output, which can be easily shown to the user. No need for further processing to generate the explanations for the users. The explanations are already part of the output model.
	The ASP ethical evaluation rules learned by the ILP learning module provide practical guidance for ethical decision making and judgment. ILP can learn effectively from small datasets, which is one of the main advantages of ILP, in addition to the comprehensibility of the generated rules by humans and machines. The learned rules are empirically valid, because the building process is tied to evidence and empirical observations.
	The ethical component can act as a governor evaluating the prospective behavior before it is executed by the agent. The outcome of the evaluation process can be used to interrupt the ongoing behavior of the agent by either prohibiting or enforcing a behavioral alternative.
	
	The system needs substantial improvements and comprehensive testing before it is ready for market. It currently presents the following challenges and limitations. Training Datasets: one of the main challenges that we have faced during this work, was the scarcity of examples. In fact, this is one of the main challenges in the ethical domain in general. This is due to two reasons. First, the field of machine ethics is a new field with very little pre-existing research work. Second, the sensitivity of the ethics domain makes it very difficult to acquire data due to privacy reasons. However, we have deployed a web application for the purpose of collecting data for building a dataset (refer Section~\ref{eval}). Another solution could be to adapt the MAS system, that we created for testing, for the creation of datasets for training.
	Limitations of the ASP translation module: the development of a more effective text-ASP translator is in our future plans. Another challenge is to fully automate the whole process: to this aim, we need to automate the generation of \textit{mode declarations} for the ILP learning module. 
	All the above mentioned limitations are subjects to our future plans. Furthermore, issues such
	as scalability and fault-tolerance are paramount to the successful operation of any application, and even more so when the application deals with something sensitive like ethics. 
	
	Building a MAS model simulating an ethical dialogue system in the domain of online customer service, helped us to get better insights into the dynamics of a corresponding real-world system, and to assess the practical challenges and limitations of building such a system. 
	We believe that the proposed architecture, and its realization in a generic agent-based platform, will facilitate the design and implementation of future ethical chatbots in different domains, maximize the reuse of existing components and the extensibility of the system to add new functionalities.

\bibliographystyle{eptcs}

\end{document}